\documentclass[conference,compsoc]{IEEEtran}

\ifCLASSOPTIONcompsoc
  \usepackage[nocompress]{cite}
\else
  \usepackage{cite}
\fi
\ifCLASSINFOpdf
\else
\fi
\usepackage{times}
\usepackage{graphicx}
\usepackage{subcaption}
\usepackage{booktabs}
\usepackage[square,sort,comma,numbers]{natbib}
\usepackage{algorithm}
\usepackage{algorithmic}
\usepackage{microtype}
\usepackage{todonotes}
\usepackage{siunitx}

\newcommand{\qnn}[2] {$\mathbf{W^{#1}A^{#2}}$}

\tikzset{
	declare function={
		mysign(\x) = (and(\x<=0, 1) * -1) +
		       (and(\x>0, 1) * 1);
		}
	}

\usepackage{hyperref}

\usepackage{glossaries}
\usepackage{algorithm}
\usepackage{multirow}

\begin{document}
%
\title{Scaling Neural Network Performance through Customized Hardware Architectures on Reconfigurable Logic}




%
\author{\IEEEauthorblockN{Michaela Blott\IEEEauthorrefmark{1},
Thomas B. Preu\ss er\IEEEauthorrefmark{1},
Nicholas Fraser\IEEEauthorrefmark{1}, 
Giulio Gambardella\IEEEauthorrefmark{1},\\
Kenneth O'Brien\IEEEauthorrefmark{1},
Yaman Umuroglu\IEEEauthorrefmark{2} and
Miriam Leeser\IEEEauthorrefmark{3}}
\IEEEauthorblockA{\IEEEauthorrefmark{1}Xilinx Research Labs, Dublin, Ireland}
\IEEEauthorblockA{\IEEEauthorrefmark{2}Norwegian University of Science and Technology, Trondheim, Norway}
\IEEEauthorblockA{\IEEEauthorrefmark{3}Northeastern University, Boston, MA, USA}}

\maketitle

\glsdisablehyper
\newacronym{FPGA}{FPGA}{Field Programmable Gate Array}
\newacronym{CNN}{CNN}{Convolutional Neural Network}
\newacronym{QNN}{QNN}{Quantized Neural Network}
\newacronym{AWS}{AWS}{Amazon Web Services}
\newacronym{BNN}{BNN}{Binarized Neural Network}
\newacronym{OFM}{OFM}{Output Feature Map}
\newacronym{IFM}{IFM}{Input Feature Map}
\newacronym{OCM}{OCM}{On-Chip Memory}
\newacronym{MVU}{MVTU}{Matrix--Vector--Threshold Unit}
\newacronym{SWU}{SWU}{Sliding Window Unit}
\newacronym{TU}{TU}{Thresholding Unit}
\newacronym{PU}{PU}{Pooling Unit}
\newacronym{II}{II}{initiation interval}
\newacronym{PE}{PE}{Processing Element}
\newacronym{NN}{NN}{Neural Network}
\newacronym{ANN}{ANN}{Artificial Neural Network}
\newacronym{FPS}{FPS}{frames per second}
\newacronym{HLS}{HLS}{High-Level Synthesis}
\newacronym{ILSVRC}{ILSVRC}{ImageNet Large Scale Visual Recognition Competition}
\newacronym{TOPS}{TOPS}{teraoperations per second}
\newacronym{GOP}{GOPS}{billion operations}
\newacronym{GFLOP}{GFLOP}{billion floating point operations}
\newacronym{LUTs}{LUT}{look up table}
\newacronym{MAC}{MAC}{multiply--accumulate}
\newacronym{RELU}{ReLU}{Rectified Linear Unit}
\newacronym{HARDTANH}{hard-tanh}{Hard Hyperbolic Tangent Function}
\begin{abstract}
Convolutional Neural Networks have dramatically improved in recent years, surpassing human accuracy on certain problems and performance exceeding that of traditional computer vision algorithms. While the compute pattern in itself is relatively simple, significant compute and memory challenges remain as CNNs may contain millions of floating-point parameters and require billions of floating-point operations to process a single image. 
These computational requirements, combined with storage footprints that exceed typical cache sizes, pose a significant performance and power challenge for modern compute architectures.
One of the promising opportunities to scale performance and power efficiency is leveraging reduced precision representations for all activations and weights as this allows to scale compute capabilities, reduce weight and feature map buffering requirements as well as energy consumption. While a small reduction in accuracy is encountered, these Quantized Neural Networks have been shown to achieve state-of-the-art accuracy on standard benchmark datasets, such as MNIST, CIFAR-10, SVHN and even ImageNet, and thus provide highly attractive design trade-offs.
Current research has focused mainly on the implementation of extreme variants with full binarization of weights and or activations, as well typically smaller input images. Within this paper, we investigate the scalability of dataflow architectures with respect to supporting various precisions for both weights and activations, larger image dimensions, and increasing numbers of feature map channels. Key contributions are a formalized approach to understanding the scalability of the existing hardware architecture with cost models and a performance prediction as a function of the target device size. We provide validating experimental results for an ImageNet classification on a server-class platform, namely the AWS F1 node. 
\end{abstract}
\section{Introduction}
\label{intro}

From speech recognition to object detection, \emph{Deep Neural Networks (DNNs)} are steadily getting better at extracting information from complex raw data.
Combined with the popularity of mobile computing and the rise of the Internet-of-Things (IoT), there is enormous potential for widespread deployment of intelligent devices, but a computational challenge remains.
A modern DNN can require billions of floating point operations to classify a single image, which is far too costly for energy-constrained compute environments, and hundreds of megabytes in memory footprint which exceed typical caching capabilities of many devices.

\begin{figure}
	\centering
	\includegraphics[height=3.5cm]{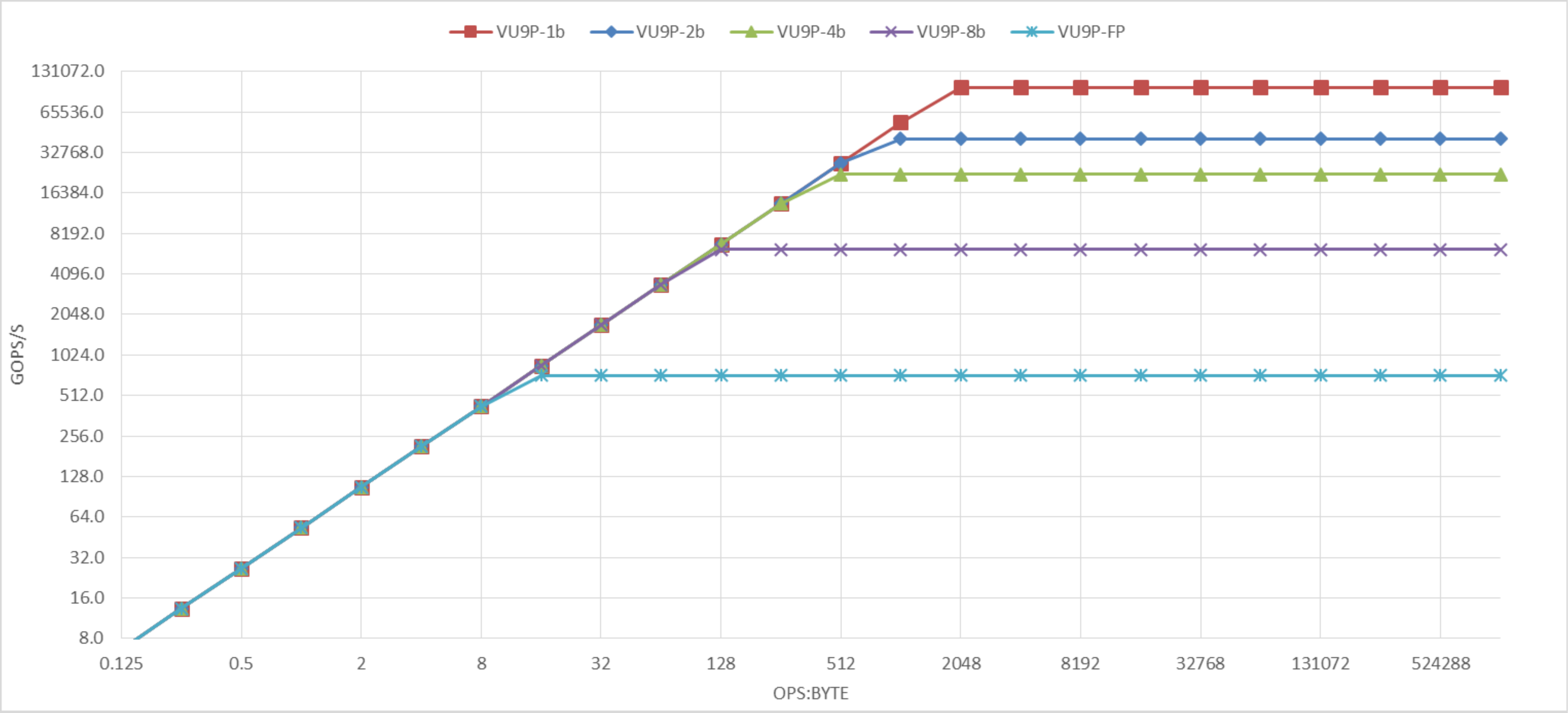}
	\caption{Theoretical Peak Performance of an AWS F1 Acceleration Card for different precisions}
	\label{fig:f1}
\end{figure}

\emph{Quantized Neural Networks (QNNs)} have recently emerged as a potential solution to this problem.
They contain convolutional, fully-connected, pooling and normalization layers similar to the floating point variants, but use a constrained set of values to represent each weight and activation in the network.
We will use the notation \qnn{w}{a} to refer to a QNN with $w$-bit weights and $a$-bit activations, and focus on cases where they represent \emph{few-bit integers} ($w, a \leq 8$).
The computational advantages of such QNNs are two-fold. First of all, each parameter and activation can be represented with a few bits, thereby a greater portion of the working set can be kept in on-chip memory, enabling greater performance, reducing off-chip memory accesses and the energy cost of data movement. Secondly QNN operations are on few-bit integers, which require only a fraction of resources and consume only a fraction of the energy \cite{energy} in a customized hardware architecture as available within an FPGA. Thereby a proportionally increased amount of operators can be instantiated in parallel, yielding a faster and more energy-efficient implementation compared to floating-point. This relationship between peak performance and precision is illustrated with the roofline model \cite{roofline} in Figure~\ref{fig:f1}.

While a quantized network will generally have reduced accuracy compared to an equivalent DNN using floating point, recent research has demonstrated significant progress in closing this accuracy gap.
Courbariaux and Hubara et al. \cite{binarynet} first demonstrated that Binarized Neural Networks (BNNs), a QNN variant with \qnn{1}{1}, could achieve competitive accuracy on smaller image recognition benchmarks like CIFAR-10 and SVHN.
XNOR-Net \cite{xnornet} improved upon this technique by adding scaling factors to better approximate the full-precision operations.
Noting that more challenging classification tasks such as ImageNet could benefit from higher-precision activations, DoReFa-Net \cite{dorefanet} used multi-bit activations and weights to further improve accuracy.
Recently, Cai et al. \cite{hwgq} proposed Half-wave Gaussian Quantization (HWGQ) to take advantage of the Gaussian-like distribution of batch-normalized activations, demonstrating \qnn{1}{2} networks with less than 5\% top-5 accuracy drop compared to floating point DNNs on the challenging ImageNet dataset, as summarized in Table \ref{tab:hwgq-accuracy}.

\begin{figure}
	\centering
	\includegraphics[width=8cm]{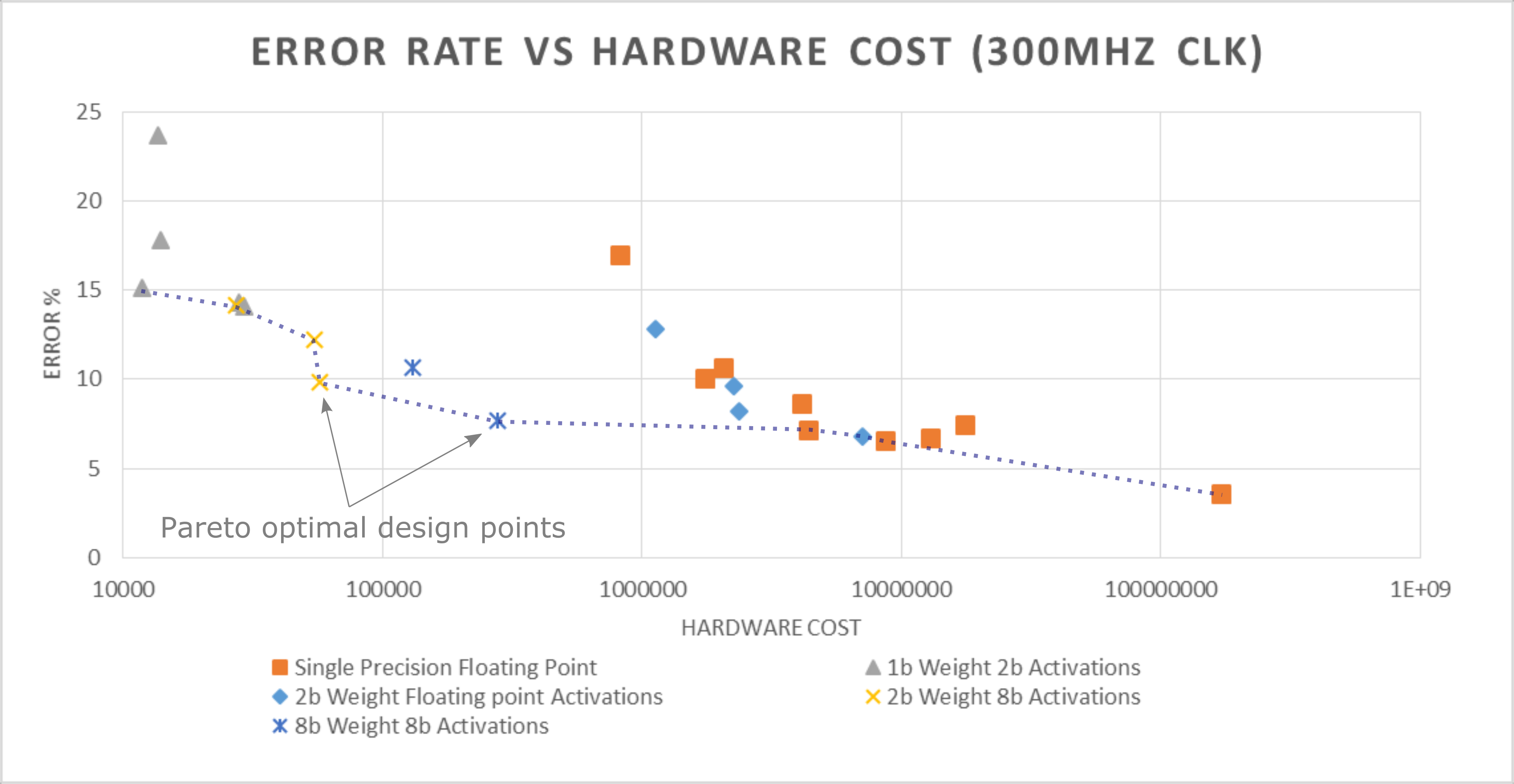}
	\caption{Design Trade-offs Accuracy vs Hardware Cost for ImageNet Classification}
	\label{fig:pareto}
\end{figure}

Given a minor degradation in accuracy cost, for many applications the combined tradeoff with associated gain in performance, efficiency and latency might provide highly attractive points within the design space.
To illustrate this, Figure~\ref{fig:pareto} depicts some data points collected from the aforementioned state of the art references as well as measured in-house, combined with our estimated hardware cost as explained in Section~\ref{performancepredictions}.
The graph illustrates achieved top-5 error rate on the y-axis against extrapolated hardware cost on the x-axis.
The best design trade-offs are the pareto-optimal points within the graph.
Given a fixed maximum error rate or conversely a maximum hardware cost, the best design compromises can be selected.

This paper investigates the scalability of dataflow architectures for originally fully binarized neural networks (BNNs) through the FINN toolflow as proposed by Umuroglu et al.~\cite{finn} and further explored in \cite{scalingbnns} in regards to increased bit precision for both weights and activations, as well as input image dimensions, and number of channels for a state of the art server class platform, namely the F1 instance in the Amazon cloud.


\begin{table}
	\caption{Accuracy of a state-of-the-art QNN \cite{hwgq}.}
	\footnotesize
	\begin{tabular}{cccc}
		\toprule
		Dataset & Network & Floating Point & \qnn{1}{2} HWGQ \cite{hwgq} \\ 
		 &  & top-1 (top-5) & top-1 (top-5) \\ 
		\midrule
		ImageNet & AlexNet & 58.5\% (81.5\%) & 52.7\% (76.3\%) \\ 
		ImageNet & GoogLeNet & 71.4\% (90.5\%) & 63.0\% (84.9\%) \\ 
		ImageNet & VGG-like & 69.8\% (89.3\%) & 64.1\% (85.6\%) \\ 
		CIFAR-10 & VGG-like & 93.2\% & 92.5\% \\
		\bottomrule
	\end{tabular} 
	\label{tab:hwgq-accuracy}
\end{table}

The rest of this paper is structured as follows: Section~\ref{impactofscaling} overviews the impact of scaling QNNs. The analysis and prediction of performances on programmable devices is described in Section~\ref{performancepredictions} together with an assessment for a real use case implemented on a VU9P FPGA in the AWS cloud.
Finally, Section~\ref{conclusions} concludes the paper.
\section{Impact of Scaling}
\label{impactofscaling}

\begin{figure}
	\centering
	\includegraphics[width=0.9\linewidth]{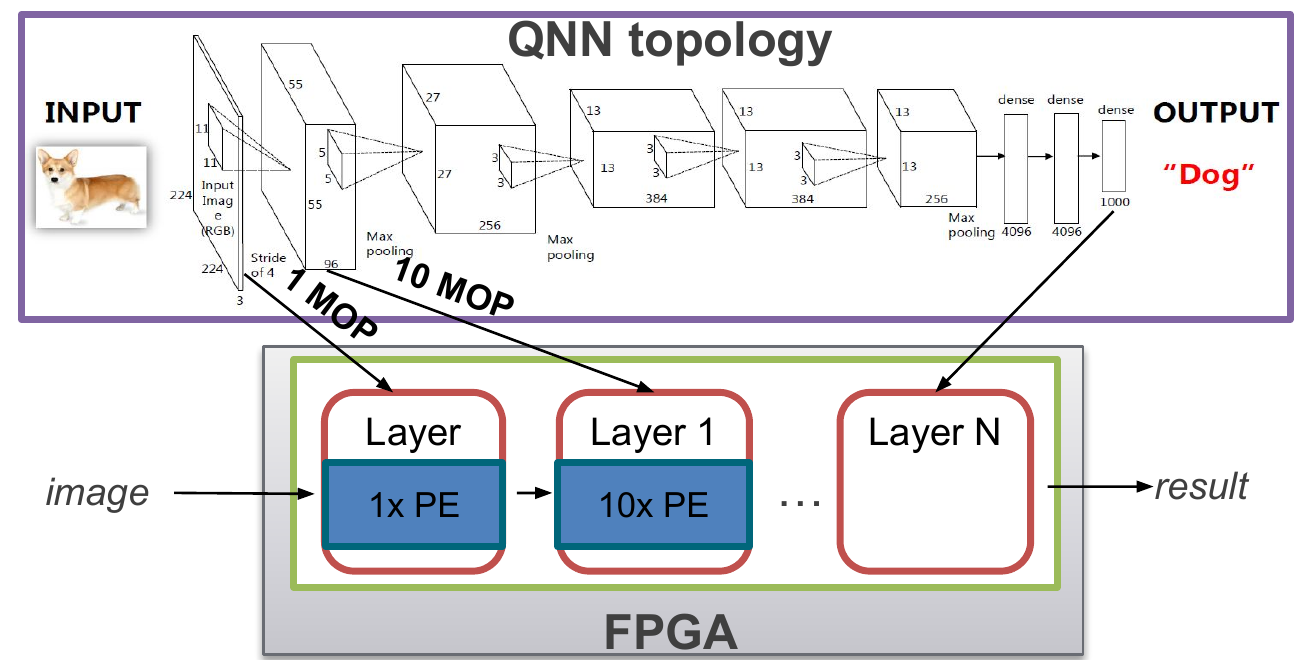}
	\caption{Heterogeneous streaming architecture.}
	\label{fig:heterogeneous-streaming}
\end{figure}

\begin{figure}
	\centering
	\includegraphics[width=0.9\linewidth]{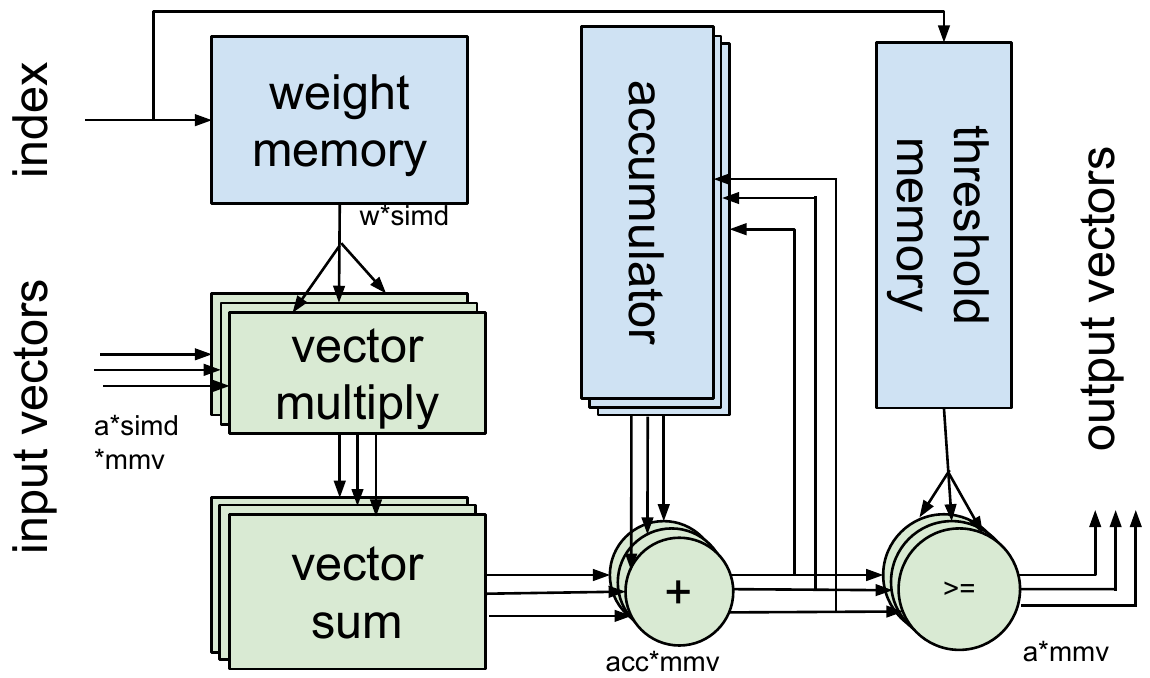}
	\caption{MMVTU datapath.}
	\label{fig:mmvtu}
\end{figure}

We adopt a heterogeneous streaming architecture as introduced by FINN. Fig.\,\ref{fig:heterogeneous-streaming} shows the key computational components: a sliding window generator buffering feature maps before a layer, weight memory, and a matrix-vector-threshold unit handling the compute.
Instead of time scheduling a single fixed datapath, we instantiate one hardware layer per QNN layer and tailor its computational resources to match the relative compute requirements of each QNN layer. We present three key improvements over FINN to enable the scaling to larger networks:\\
\textbf{Higher-precision operators.} While FINN supports binarized weights and activations, our architecture adds full per-layer customization of the precisions of weights, input and output activations and the accumulators for the dot product. These customizable operators are implemented as templated functions in C++ using Vivado HLS. They are mapped to LUTs for narrow bit widths and to DSPs for 8-bit operations.\\
\textbf{Improved Matrix Vector-Threshold Units.} We enhanced the originally proposed computational cores, the Matrix-Vector-Threshold Units, to process multiple vectors from different input feature maps in parallel. This new MMVTU shown in Fig.\ref{fig:mmvtu} re-uses convolutional weights across all parallel computations. This improvement enables a performance scaling with an increased BRAM utilization.\\
\textbf{Improved Sliding Window Unit (SWU).} We introduced a new SWU which has reduced buffer size requirements. This was achieved by buffering the absolute bare minimum amount of data before the next MVTU can start computation. The minimum equates to a few consecutive rows as the height of the convolutional kernel must be kept available. For elasticity reasons, an extra row buffer is used to collect new incoming image data.

\section{Performance Prediction and Evaluation}
\label{performancepredictions}
Different applications have different accuracy, performance and latency requirements. Combined with the parameterizable architecture, this yields a design space too vast to explore by synthesizing each design point. The quick navigation of this design space rather requires a tool that is able to identify the interesting design options automatically.

The tool is driven by analytic resource and performance models. Computed design points are constrained by the resources available on the particular selected device. Assuming a minimal non-parallelized implementation of the desired network topology, it is able to determine if the chosen device is at all suitable for an implementation. If this is the case, balanced scaling of the compute resources in those layers, which currently constitute the computational bottleneck pushes the performance to the desired level, at least, as much as the available device resources allow.

The computational concurrency of a convolutional layer is controlled by three parameters: (a) the neuron count $PE$, i.e. the number of processing elements concurrently working on distinct output channels, (b) the $SIMD$ count, i.e. the number of input channels processed within one clock cycle, and (c) the multi-vector count capturing the concurrent duplication of this compute structure across multiple input images. As explained in the MMVTU section, the motivation of adding this third degree of parallelization is the immediate reuse of kernel weights for independent input images for a more economical use of on-chip memory bandwidth. However, this coarse-grain parallelization is not able to reduce compute latency but can only increase the overall throughput.

\begin{table}
\caption{Convolutional Layer Parameters}\label{tabParams}
\centerline{\begin{tabular}{@{}ll@{}}\toprule
   $M$            & parallel vectors processed by MMVTU\\
   $N, C$         & input feature map width and channels\\
   $K\times K, S$ & kernel dimension and stride\\
   $C'$           & output feature map channels\\
   $A, W$         & bit width (precision) of activations / weights\\\bottomrule
\end{tabular}}
\end{table}

The BRAM requirements of a convolutional layer, as characterized by the parameters of Tab.\,\ref{tabParams}, comprise the line buffers of its sliding window unit and the storage for the convolutional weights.
The corresponding numbers of BRAM modules can be directly derived from the implemented memory layout. The line buffer occupies as many BRAM modules
as specified by Eq.\,(\ref{eqSWURAM}).
\begin{align}
	\label{eqSWURAM}
	BRAM_{swu} &= M\left(\left\lceil\frac{K}{S}\right\rceil+1\right)
	\left\lceil\frac{S\cdot N}{512}\right\rceil\times
    \left\lceil\frac{C\cdot A}{36}\right\rceil
\end{align}

The multi-vector count scales linearly on the highest level of the equation. Otherwise, independent stripes of memory are
used for each set of rows that can be released independently once the whole width of a line has been processed. An additional
memory stripe is used as assembly buffer for the new image data coming in. This accounts for the first parenthesized factor.
The remaining two factors capture the depth and the width of the memory stripes, which are potentially fragmented due to the depth and word width of the builtin BRAM modules.

\begin{align}
	\label{eqWEIGHTRAM}
	BRAM_{weights} &= PE\left\lceil\frac{WM\cdot 36}{512}\right\rceil\times
	\left\lceil\frac{SIMD\cdot W}{36}\right\rceil\\\nonumber
	&\mbox{with}\quad WM = \frac{K^2\cdot C \cdot C'}{SIMD\cdot PE}
\end{align}

Similarly, Eq.\,(\ref{eqWEIGHTRAM}) captures the number of BRAM modules needed to implement the weight memory of a convolutional layer. Its overall size is determined by the product of the squared kernel dimension and the numbers of input as well as output feature map channels. This memory volume is split into separate memories, one for each processing element. The parallel access of SIMD weights determines the word width used by the implementation. Again, memory depth and word size may be fragmented by the physical dimensions of the available BRAM modules.

The actual computation is bounded by the availability of LUT resources. The complexity of an individual MAC operation is scaled with all three dimensions of concurrency as shown in Eq.\,(\ref{eqLUTS}). The elementary LUT cost of a single product has been determined empirically and validated from a series of HLS synthesis runs, the targeted synthesis flow. We expect to reduce these elementary costs by adopting more optimal summations in the computation of the MAC operations as suggested by Preu{\ss}er \cite{preusser:2017}.
\begin{align}
\label{eqLUTS}
LUTs &= M\cdot PE\cdot SIMD\cdot f\left(A,W\right)
\end{align}

Given this cost model, we can extrapolate the maximum possible performance for our example network within the constraints of a given device.
As can be seen from the table, the performance decreases hyperbolically with number of activation bits.
However, the increase from  w-bit weights from 1 to 2 is almost negligible as the binarized version is in reality a bipolar representation where the two possible values are -1 and +1, which require 2 bits in representation.

\begin{figure}
	\centering
	\includegraphics[height=3.5cm]{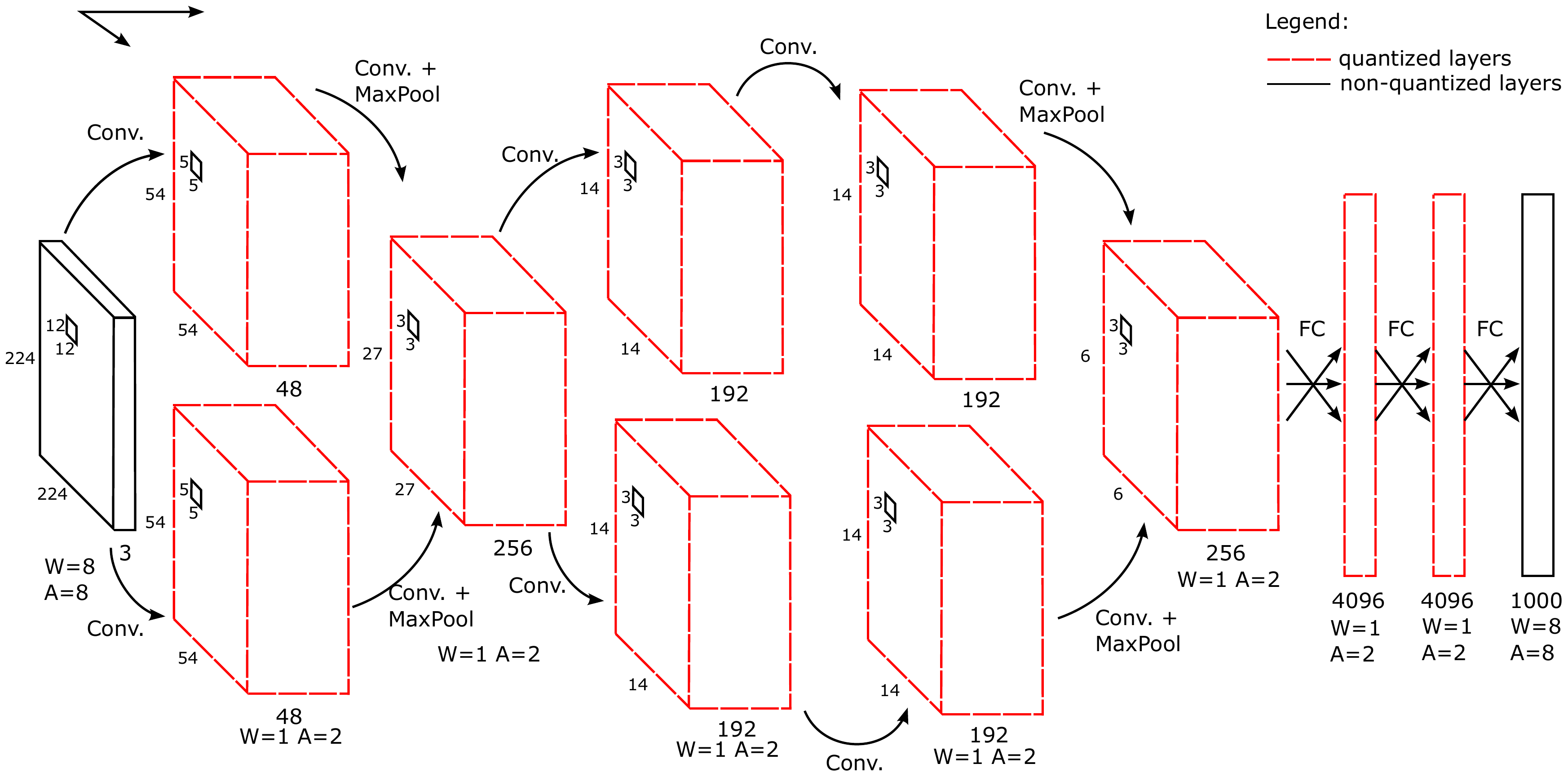}
	\caption{DoReFa-Net topology}
	\label{fig:dorefanet}
\end{figure}

For our experimental setup, we selected DoReFa-Net~\cite{dorefanet}, which we trained using tensorpack\footnote{https://github.com/ppwwyyxx/tensorpack} and is depicted in Figure~\ref{fig:dorefanet}.
Binarised weights are trained the same way as in DoReFa-Net~\cite{dorefanet}, with binary values being used as the mean of the underlying real weight values in each layer.
For w-bit weights ($w > 1$), we use a 2's compliment representation with a fractional length of $w-2$.
For a-bit activations, we implement a clipped ReLU function, $f(x)$ as follows:
\begin{align}
    f(x) = 
    \begin{cases}
    x,& \mathrm{if\ } 0 \leq x \leq 1\\
    0 & \mathrm{otherwise}
    \end{cases}
\end{align}
These values are then quantized to $n$ equal spaced values across this range, where $n = 2^a$.

Table~\ref{tab:dorefa-accuracy} summarizes the achieved accuracy results for Top-5 ImageNet classification together with our projected performance given the previously introduced formula,
assuming 80\% device utilization, and 250MHz clock frequency.
In Table~\ref{tab:dorefa-accuracy}, \qnn{w}{a} represents a bit-depth of $w$ and $a$ for the weights and activations respectively,
Similar to previous works, all DoReFa-Net configurations use a higher bit depth for weights in the first and last layers.
In this work, 8-bits weights are assumed for these layers.

\begin{table}
\centering
\caption{Single crop validation error vs Performance of DoReFa-Net on ImageNet and multiple precisions}
\begin{minipage}{\linewidth}
\begin{center}
\begin{tabular}{lrrrr}
\toprule
Bit- & Top-5    & Top-1    & kFPS & kFPS\\
depth& Err (\%) & Err (\%) & (Est.) & (Ach.)\\
\midrule
\qnn{1}{1}  & 30.9 & 54.6 & 15.4&\\
\qnn{1}{2}  & 26.9 & 50.7 & 8.5&\\
\qnn{1}{2}$^\dagger$ & 26.0 & 49.7 & 16.4& 3.94\\
\qnn{2}{2}  & 29.4 & 53.4 & 7.6&\\
\qnn{4}{4}  & 24.2 & 47.5 & 4.1&\\
\qnn{8}{8}  & 22.8 & 46.6 & 1.4&\\
\bottomrule
\end{tabular}
\end{center}
$^\dagger$~\parbox[t]{.9\linewidth}{\footnotesize{A smaller, retrained DoReFa-Net with approximately 50\% of the operations of the other versions.}}
\end{minipage}
\vspace{-2ex}
\label{tab:dorefa-accuracy}
\end{table}


We evaluated our cost function and performance predictions for the pruned variant of DoReFa-Net implementation \qnn{1}{2}.
We measure the performance excluding the overhead of the data transfer from and to the host and measured 3915 fps at a clock frequency of 109MHz with a latency of 0.432msec.
While the actual measured results shows an obvious degradation in performance, the discrepancies are easily explained. Firstly, the actual prototype currently clocks at 109MHz, which is substantially below the estimates which assumed 400MHz and what is achievable in hardware with more effort spent in floorplanning and timing closure. Taking this into account, we modify our estimate to 4.1kfps which is within 5\% of the measured results and demonstrate the hyperbolic nature of scaling bit-precisions in activation functions. Remaining discrepancies are caused by resource fragmentation due to the integral scaling of the compute for each layer and stalling due to insufficient buffer sizing between layers. We believe for further accuracy in prediction, this requires a dynamic modelling effort in the future.

\section{Conclusions}
\label{conclusions}
Within this paper, we investigate the scalability of dataflow architectures which were introduced in conjunction with binarized neural networks, with respect to supporting increasing precisions for both weights and activations, larger image dimensions, and increasing numbers of feature map channels. Key contributions are a formalized approach to understanding the scalability of the existing hardware components with cost models and the resulting capability for performance prediction as a function of the target device size and bit-precision in leveraged datatypes. We provide validating experimental results for an ImageNet classification on the AWS's F1 node. Future work will include the refinement of the cost model.

\bibliography{main.bib}
\bibliographystyle{IEEEtran}


\end{document}